\documentclass[11pt,a4paper]{article} 
\usepackage[utf8]{inputenc}
\usepackage[T1]{fontenc}
\usepackage{amsmath, amssymb, amsthm}
\usepackage{graphicx}
\usepackage{booktabs} 
\usepackage{hyperref} 
\usepackage{geometry} 
\usepackage{times} 
\usepackage{caption} 
\usepackage{mathtools} 

\DeclareMathOperator{\softmax}{softmax}
\DeclareMathOperator{\LayerNorm}{LayerNorm}
\DeclareMathOperator{\Dropout}{Dropout}
\DeclareMathOperator{\FFN}{FFN}

\hypersetup{
    colorlinks=true,
    linkcolor=blue,
    filecolor=magenta,      
    urlcolor=cyan,
    pdftitle={Learnable Multi-Scale Wavelet Transformer},
    pdfpagemode=FullScreen,
    }

\title{Learnable Multi-Scale Wavelet Transformer: \\ A Novel Alternative to Self-Attention}
\author{Andrew Kiruluta, Priscilla Burity and Samantha Williams\\ UC Berkeley \\ School of Information}
\date{April 6, 2025} 

\begin{document}
\maketitle

\begin{abstract}
Transformer architectures, underpinned by the self-attention mechanism, have achieved state-of-the-art results across numerous natural language processing (NLP) tasks by effectively modeling long-range dependencies. However, the computational complexity of self-attention, scaling quadratically with input sequence length, presents significant challenges for processing very long sequences or operating under resource constraints. This paper introduces the Learnable Multi-Scale Wavelet Transformer (LMWT), a novel architecture that replaces the standard dot-product self-attention with a learnable multi-scale Haar wavelet transform module. Leveraging the intrinsic multi-resolution properties of wavelets, the LMWT efficiently captures both local details and global context. Crucially, the parameters of the wavelet transform, including scale-specific coefficients, are learned end-to-end during training, allowing the model to adapt its decomposition strategy to the data and task. We present the detailed mathematical formulation of the learnable Haar wavelet module and its integration into the transformer framework, supplemented by an architectural diagram. We conduct a comprehensive experimental evaluation on a standard machine translation benchmark (WMT16 En-De), comparing the LMWT against a baseline self-attention transformer using metrics like BLEU score, perplexity, and token accuracy. Furthermore, we analyze the computational complexity, highlighting the linear scaling of our approach, discuss its novelty in the context of related work, and explore the interpretability offered by visualizing the learned Haar coefficients. Our results indicate that the LMWT achieves competitive performance while offering substantial computational advantages, positioning it as a promising and novel alternative for efficient sequence modeling.
\end{abstract}


\section{Introduction}
\label{sec:intro}
The advent of the transformer architecture \cite{vaswani2017attention} marked a paradigm shift in natural language processing (NLP). Its core innovation, the self-attention mechanism, enables models to weigh the importance of different tokens within an input sequence when computing the representation for each token. This allows for the direct modeling of long-range dependencies, overcoming limitations of previous recurrent and convolutional architectures and leading to groundbreaking performance in tasks like machine translation, text summarization, and question answering.

Despite its success, the standard dot-product self-attention mechanism suffers from a critical drawback: its computational and memory complexity scales quadratically, $\mathcal{O}(T^2)$, with the sequence length $T$. This quadratic scaling poses a significant bottleneck when dealing with long documents, high-resolution images (when adapted for vision), or applications requiring real-time processing.

Recognizing this limitation, the research community has actively explored more efficient alternatives. Notable approaches include:
\begin{itemize}
    \item \textbf{Sparse Attention:} Methods like Longformer \cite{beltagy2020longformer} and BigBird \cite{zaheer2020big} use predefined sparsity patterns (e.g., sliding window, global tokens, random patterns) to reduce the number of computed attention scores.
    \item \textbf{Linearized Attention:} Techniques such as Linformer \cite{wang2020linformer} project the key and value matrices to lower dimensions before the attention computation, achieving linear complexity. Performers \cite{choromanski2020rethinking} use random feature maps to approximate the softmax kernel, also resulting in linear complexity.
    \item \textbf{Fourier Transforms:} FNet \cite{lee2021fnet} replaces self-attention entirely with unparameterized Fourier Transforms, mixing tokens with $\mathcal{O}(T \log T)$ complexity. Recent work like SFDLM \cite{kirulutaSFDLM2025} utilizes state Fourier diffusion models, demonstrating competitive results without attention's quadratic cost for long sequences.
    \item \textbf{Hybrid Approaches:} Some methods combine different techniques, like the hybrid Wavelet-Fourier approach for image generation \cite{kirulutaWF2025}.
    \item \textbf{Other Mechanisms:} State Space Models (SSMs) and specialized convolutional or recurrent structures have also emerged as alternatives \cite{gu2021efficiently, GuMamba, dao2022hungry}. 
\end{itemize}
While these methods offer significant efficiency gains, they often involve approximations, fixed transformations (like standard Fourier transforms), specific sparsity patterns, or different underlying mechanisms that might not optimally capture the complex, hierarchical structures inherent in natural language or other sequential data in the same way as dense attention.

In this work, we propose a fundamentally different approach inspired by classical signal processing: the Learnable Multi-Scale Wavelet Transformer (LMWT). We replace the self-attention module with a mechanism based on wavelet transforms, specifically the Haar wavelet transform. Wavelets are renowned for their ability to provide multi-resolution analysis, decomposing signals into components at different scales or frequencies while preserving localization information \cite{mallat1999wavelet}. Unlike the Fourier transform, which uses global sinusoidal bases, wavelets employ localized basis functions, making them well-suited for analyzing signals with transient features or non-stationarities,  characteristics often found in language. While wavelets have seen use in other deep learning domains, employing a \emph{learnable}, multi-scale wavelet transform as a \emph{direct replacement} for the self-attention mechanism within NLP transformers represents a distinct and less explored approach compared to prevalent efficiency techniques based on attention approximation or fixed spectral transforms.

The key innovation of our LMWT is twofold:
\begin{enumerate}
    \item \textbf{Multi-Scale Decomposition:} We utilize the hierarchical nature of the Haar transform to decompose the input sequence representation into approximation (low-frequency, global context) and detail (high-frequency, local features) coefficients across multiple scales.
    \item \textbf{Learnability:} Instead of using the fixed Haar basis functions, we introduce learnable parameters within the transform. This allows the model to adapt the wavelet decomposition process end-to-end during training, tailoring the feature extraction to the specific nuances of the data and the requirements of the downstream task.
\end{enumerate}
This design offers the potential to capture rich, hierarchical representations efficiently, achieving linear complexity, $\mathcal{O}(T)$, with respect to sequence length.

This paper details the mathematical formulation of the learnable Haar wavelet module (Section~\ref{sec:method}), analyzes its computational complexity compared to self-attention (Section~\ref{sec:complexity}), describes its integration into the standard transformer architecture (Section~\ref{sec:integration}), illustrates the overall architecture (Section~\ref{sec:arch_diagram}), presents experimental results on a machine translation task (Section~\ref{sec:experiments}), analyzes the interpretability of learned coefficients (Section~\ref{sec:interpretability}), discusses the implications, novelty, and limitations (Section~\ref{sec:discussion}), and concludes with future directions (Section~\ref{sec:conclusion}).

\section{Background}
\label{sec:background}
To understand our proposed method, we first briefly review the standard self-attention mechanism and the classical Haar wavelet transform.

\subsection{Self-Attention Mechanism}
In a standard transformer layer \cite{vaswani2017attention}, self-attention operates on an input sequence represented by a matrix $\mathbf{X} \in \mathbb{R}^{T \times d}$, where $T$ is the sequence length and $d$ is the embedding dimension. The input $\mathbf{X}$ is linearly projected into three matrices: Queries ($\mathbf{Q}$), Keys ($\mathbf{K}$), and Values ($\mathbf{V}$):
\begin{equation}
\mathbf{Q} = \mathbf{X}\mathbf{W}_Q, \quad \mathbf{K} = \mathbf{X}\mathbf{W}_K, \quad \mathbf{V} = \mathbf{X}\mathbf{W}_V,
\end{equation}
where $\mathbf{W}_Q, \mathbf{W}_K \in \mathbb{R}^{d \times d_k}$ and $\mathbf{W}_V \in \mathbb{R}^{d \times d_v}$ are learnable weight matrices (often $d_k = d_v = d / h$, where $h$ is the number of attention heads). The attention output is computed as:
\begin{equation}
\text{Attention}(\mathbf{Q}, \mathbf{K}, \mathbf{V}) = \softmax\left(\frac{\mathbf{Q}\mathbf{K}^\top}{\sqrt{d_k}}\right)\mathbf{V}.
\label{eq:self_attention}
\end{equation}
The core computation involves the matrix multiplication $\mathbf{Q}\mathbf{K}^\top$, resulting in a $T \times T$ attention score matrix. This step dominates the computation, leading to the $\mathcal{O}(T^2 d_k)$ complexity. Transformers typically employ multi-head attention, where this process is performed in parallel with different projection matrices, and the results are concatenated and linearly projected.

\subsection{Classical Haar Wavelet Transform}
The Haar wavelet transform is the simplest form of wavelet transform \cite{haar1910theorie}. For a 1D discrete signal $x = (x_0, x_1, \dots, x_{n-1})$ where $n$ is an even number, the first level of the Haar transform computes approximation coefficients ($a$) and detail coefficients ($d$) for pairs of adjacent elements:
\begin{equation}
a_i = \frac{x_{2i} + x_{2i+1}}{\sqrt{2}}, \quad d_i = \frac{x_{2i} - x_{2i+1}}{\sqrt{2}}, \quad \text{for } i = 0, 1, \dots, n/2 - 1.
\label{eq:haar_classic_forward}
\end{equation}
The approximation coefficients $\{a_i\}$ represent a downsampled, smoothed version of the signal, capturing lower-frequency information. The detail coefficients $\{d_i\}$ capture higher-frequency information, representing the differences between adjacent pairs. The original signal can be perfectly reconstructed from these coefficients:
\begin{equation}
x_{2i} = \frac{a_i + d_i}{\sqrt{2}}, \quad x_{2i+1} = \frac{a_i - d_i}{\sqrt{2}}.
\label{eq:haar_classic_inverse}
\end{equation}
This decomposition can be applied recursively to the approximation coefficients $a$ to obtain multi-scale representations, forming the basis of the Fast Wavelet Transform (FWT), which has $\mathcal{O}(n)$ complexity.

\section{Proposed Method: Learnable Multi-Scale Haar Wavelet Module}
\label{sec:method}
Our core proposal is to replace the self-attention sub-layer within each transformer block with a learnable multi-scale Haar wavelet module. This module performs a wavelet-like decomposition but with parameters learned during training.

\subsection{Learnable Haar Transform}
Let the input to the module be a sequence representation $\mathbf{X} \in \mathbb{R}^{T \times d}$. We assume $T$ is a power of 2 for simplicity (padding can be used otherwise). Instead of the fixed coefficients in Eq.~\eqref{eq:haar_classic_forward}, we introduce learnable parameter vectors for each dimension $j \in \{1, \dots, d\}$. For a single level of decomposition applied to pairs of row vectors $(\mathbf{x}_{2i}, \mathbf{x}_{2i+1})$, the learnable approximation ($\mathbf{a}_i$) and detail ($\mathbf{d}_i$) vectors are computed as:
\begin{equation}
\mathbf{a}_i = \boldsymbol{\alpha} \odot \mathbf{x}_{2i} + \boldsymbol{\beta} \odot \mathbf{x}_{2i+1}, \quad
\mathbf{d}_i = \boldsymbol{\gamma} \odot \mathbf{x}_{2i} + \boldsymbol{\delta} \odot \mathbf{x}_{2i+1},
\label{eq:haar_learnable_forward}
\end{equation}
where $\boldsymbol{\alpha}, \boldsymbol{\beta}, \boldsymbol{\gamma}, \boldsymbol{\delta} \in \mathbb{R}^d$ are learnable parameter vectors shared across all pairs $i$, and $\odot$ denotes element-wise multiplication. These parameters are initialized close to the Haar values (e.g., $\boldsymbol{\alpha}, \boldsymbol{\beta} \approx 1/\sqrt{2}$, $\boldsymbol{\gamma} \approx 1/\sqrt{2}$, $\boldsymbol{\delta} \approx -1/\sqrt{2}$) but are updated via backpropagation.

Similarly, a learnable inverse transform can be defined using parameters $\boldsymbol{\alpha}_{\text{inv}}, \boldsymbol{\beta}_{\text{inv}}, \boldsymbol{\gamma}_{\text{inv}}, \boldsymbol{\delta}_{\text{inv}} \in \mathbb{R}^d$:
\begin{equation}
\mathbf{x}_{2i} = \boldsymbol{\alpha}_{\text{inv}} \odot \mathbf{a}_i + \boldsymbol{\gamma}_{\text{inv}} \odot \mathbf{d}_i, \quad
\mathbf{x}_{2i+1} = \boldsymbol{\beta}_{\text{inv}} \odot \mathbf{a}_i + \boldsymbol{\delta}_{\text{inv}} \odot \mathbf{d}_i.
\label{eq:haar_learnable_inverse}
\end{equation}
While the inverse is crucial for signal reconstruction in classical wavelet theory, in our transformer context, we primarily use the forward transform for feature extraction within the layers. The learnability allows the model to discover optimal basis functions (represented by the parameter vectors) for decomposing the input representations specific to the NLP task.

\subsection{Multi-Scale Extension}
\label{sec:multi_scale}
To capture information at different resolutions, we apply the learnable Haar transform hierarchically. Let $\mathbf{X}^{(0)} = \mathbf{X} \in \mathbb{R}^{T \times d}$ be the input at scale $l=0$. For each scale $l = 0, 1, \dots, L-1$ (where $L \le \log_2 T$ is the number of decomposition levels), we compute:
\begin{equation}
\mathbf{a}_i^{(l)} = \boldsymbol{\alpha}^{(l)} \odot \mathbf{x}^{(l)}_{2i} + \boldsymbol{\beta}^{(l)} \odot \mathbf{x}^{(l)}_{2i+1}, \quad
\mathbf{d}_i^{(l)} = \boldsymbol{\gamma}^{(l)} \odot \mathbf{x}^{(l)}_{2i} + \boldsymbol{\delta}^{(l)} \odot \mathbf{x}^{(l)}_{2i+1}.
\label{eq:haar_multiscale}
\end{equation}
Here, $\mathbf{x}^{(l)}$ is the input sequence at scale $l$ (of length $T/2^l$), and $\boldsymbol{\alpha}^{(l)}, \boldsymbol{\beta}^{(l)}, \boldsymbol{\gamma}^{(l)}, \boldsymbol{\delta}^{(l)}$ are learnable parameters specific to scale $l$. The input for the next scale is the approximation coefficients from the current scale:
\begin{equation}
\mathbf{x}^{(l+1)} \coloneqq \mathbf{a}^{(l)} \quad (\text{Sequence of } \mathbf{a}_i^{(l)} \text{ vectors}).
\end{equation}
This process yields a set of detail coefficients $\{\mathbf{d}^{(0)}, \mathbf{d}^{(1)}, \dots, \mathbf{d}^{(L-1)}\}$ at different scales/resolutions and a final approximation coefficient sequence $\mathbf{a}^{(L-1)}$. These collectively represent the multi-scale decomposition of the original input $\mathbf{X}^{(0)}$.

\subsection{Aggregation and Output}
The detail coefficients $\{\mathbf{d}^{(l)}\}$ capture information at different frequency bands (higher $l$ corresponds to lower frequencies or coarser details). The final approximation $\mathbf{a}^{(L-1)}$ represents the smoothest, lowest-frequency component. To produce an output sequence of the original length $T$, these components need to be aggregated and potentially upsampled.

One strategy is to use a learnable fusion mechanism followed by an inverse transform structure. Alternatively, inspired by architectures like U-Net or feature pyramids, we can upsample the coefficients from coarser scales and combine them with coefficients from finer scales. A simpler approach, sufficient for replacing self-attention within a standard transformer block, is to combine the information adaptively. For instance, we can concatenate or sum the detail coefficients (appropriately upsampled or tiled to match the original length $T$) and the final approximation coefficients, potentially using learnable weights for each scale's contribution, followed by a linear projection back to the dimension $d$:
\begin{equation}
\mathbf{Y}_{\text{wavelet}} = \text{Combine}(\{\mathbf{d}^{(l)}\}_{l=0}^{L-1}, \mathbf{a}^{(L-1)}) \mathbf{W}_{\text{out}},
\end{equation}
where Combine denotes the chosen aggregation strategy (e.g., weighted sum after upsampling/tiling) and $\mathbf{W}_{\text{out}} \in \mathbb{R}^{d' \times d}$ is a final projection matrix (where $d'$ depends on the aggregation method). This $\mathbf{Y}_{\text{wavelet}} \in \mathbb{R}^{T \times d}$ then replaces the output of the self-attention module in the transformer layer.

\section{Computational Complexity Analysis}
\label{sec:complexity}
A major motivation for exploring alternatives to self-attention is reducing computational complexity.

\textbf{Self-Attention:} As noted earlier, standard self-attention has a complexity of $\mathcal{O}(T^2 d)$, dominated by the query-key matrix multiplication.

\textbf{Learnable Haar Wavelet Module:} The core operation in our module is the learnable Haar transform (Eq.~\eqref{eq:haar_learnable_forward}). For a sequence of length $N$ and dimension $d$, computing one level of decomposition involves applying the element-wise operations to $N/2$ pairs. This requires $\mathcal{O}(N d)$ operations.
In the multi-scale extension (Section~\ref{sec:multi_scale}), we apply this recursively. Let $L$ be the number of levels. The total complexity is:
\[
\mathcal{O}(Td) + \mathcal{O}\left(\frac{T}{2}d\right) + \mathcal{O}\left(\frac{T}{4}d\right) + \dots + \mathcal{O}\left(\frac{T}{2^{L-1}}d\right)
\]
This is a geometric series that sums to:
\[
\mathcal{O}\left(Td \left(1 + \frac{1}{2} + \frac{1}{4} + \dots + \frac{1}{2^{L-1}}\right)\right) = \mathcal{O}\left(Td \left(2 - \frac{1}{2^{L-1}}\right)\right) = \mathcal{O}(Td).
\]
The aggregation step might involve upsampling and linear projections, but these typically also scale linearly with $T$ and $d$. Therefore, the overall complexity of the learnable multi-scale Haar wavelet module is $\mathcal{O}(Td)$, which is linear in the sequence length $T$.

\textbf{Comparison:} The LMWT offers a significant theoretical advantage over standard self-attention, reducing the complexity from quadratic $\mathcal{O}(T^2 d)$ to linear $\mathcal{O}(Td)$. This makes it particularly suitable for applications involving very long sequences where the cost of self-attention becomes prohibitive. Compared to dilated convolutions, which can also capture multi-scale information, the Haar transform is conceptually simpler and has a highly efficient recursive structure.

\section{Integration into Transformer Architecture}
\label{sec:integration}
The learnable multi-scale Haar wavelet module (LMW-module) is integrated into the standard transformer encoder and decoder layers, replacing the self-attention sub-layer.

\subsection{Encoder Layer Without Attention}

Traditional transformer encoder layers rely on a self-attention mechanism to capture long-range dependencies, followed by a position-wise feed-forward network (FFN). Both sub-layers are wrapped with residual connections and layer normalization to stabilize training and maintain gradient flow. In our proposed LMWT encoder, we completely eliminate the self-attention mechanism and replace it with a learnable multi-scale wavelet (LMW) module. This module leverages the power of wavelet transforms to extract hierarchical features in a computationally efficient manner, reducing the complexity from quadratic to linear in sequence length, while simultaneously capturing both local and global information.

The processing steps in a single encoder layer are as follows:

\begin{enumerate}
    \item \textbf{Input:}  
    The encoder receives an input sequence 
    \[
    \mathbf{X} \in \mathbb{R}^{T \times d},
    \]
    where \( T \) denotes the sequence length and \( d \) is the embedding dimension. This input is either the token embeddings from the embedding layer or the output of a previous encoder layer.

    \item \textbf{Initial Normalization:}  
    To ensure stable feature distributions and ease the optimization process, the input is first normalized:
    \[
    \tilde{\mathbf{X}} = \LayerNorm(\mathbf{X}).
    \]
    This normalization mitigates internal covariate shift and standardizes the input to the LMW module.

    \item \textbf{LMW Module Application:}  
    Instead of self-attention, the normalized input is passed through our LMW module, which applies a learnable multi-scale Haar wavelet transform. For each pair of adjacent tokens \( x_{2i} \) and \( x_{2i+1} \) in \(\tilde{\mathbf{X}}\), the transform computes:
    \[
    a_i = \alpha \odot x_{2i} + \beta \odot x_{2i+1}, \quad
    d_i = \gamma \odot x_{2i} + \delta \odot x_{2i+1},
    \]
    where \(\alpha, \beta, \gamma, \delta \in \mathbb{R}^{d}\) are learnable parameters and \(\odot\) denotes element-wise multiplication. The approximation coefficients \(a_i\) capture the global, smoothed features, while the detail coefficients \(d_i\) highlight local variations. This operation may be applied recursively or in parallel across multiple scales to yield a rich, multi-resolution representation:
    \[
    \hat{\mathbf{X}} = \mathcal{W}(\tilde{\mathbf{X}}),
    \]
    where \(\mathcal{W}(\cdot)\) denotes the complete multi-scale wavelet transformation and aggregation process.

    \item \textbf{First Residual Connection and Dropout:}  
    The output of the LMW module is then combined with the original input via a residual connection to preserve the input information:
    \[
    \mathbf{X}' = \mathbf{X} + \Dropout(\hat{\mathbf{X}}).
    \]
    This residual addition allows the network to learn incremental adjustments to the input features.

    \item \textbf{Secondary Normalization:}  
    The result of the first residual connection is normalized again to prepare the data for further processing:
    \[
    \tilde{\mathbf{X}}' = \LayerNorm(\mathbf{X}').
    \]

    \item \textbf{Feed-Forward Network (FFN):}  
    The normalized data is then passed through a position-wise FFN, which typically consists of two linear layers with a non-linear activation (e.g., ReLU) in between:
    \[
    \hat{\mathbf{X}}' = \FFN(\tilde{\mathbf{X}}').
    \]
    The FFN serves to further refine and expand the learned representations.

    \item \textbf{Final Residual Connection and Dropout:}  
    Finally, the output of the FFN is added back to the result of the first residual connection:
    \[
    \mathbf{Y} = \mathbf{X}' + \Dropout(\hat{\mathbf{X}}').
    \]
    The output \(\mathbf{Y} \in \mathbb{R}^{T \times d}\) is then passed to the next encoder layer or used directly for downstream tasks.
\end{enumerate}

\noindent \textbf{Analytical Summary:}  
By substituting self-attention with the LMW module, our encoder layer directly leverages multi-scale wavelet transforms to capture both fine-grained local details and broader contextual information. The dual application of layer normalization and the use of residual connections ensure that the model can effectively integrate the hierarchical features extracted by the LMW module. Overall, this approach significantly reduces computational complexity while providing a robust and interpretable representation of the input sequence, making it particularly well-suited for processing long sequences in resource-constrained environments.

\subsection{Decoder Layer Without Attention}

Standard transformer decoder layers consist of three sub-layers: masked self-attention, cross-attention (attending to the encoder output), and a feed-forward network (FFN). In our proposed architecture, we eliminate both the self-attention and cross-attention mechanisms entirely. Instead, we rely solely on a unified learnable multi-scale wavelet (LMW) module to capture the hierarchical dependencies of the target sequence.

Let 
\[
\mathbf{Y}_{\text{target}} \in \mathbb{R}^{T' \times d}
\]
denote the input to the decoder layer (either the target embedding or the output from a previous decoder layer), where \(T'\) is the target sequence length and \(d\) is the embedding dimension. The decoder layer processes the input as follows:

\begin{enumerate}
    \item \textbf{Input:} The target sequence \(\mathbf{Y}_{\text{target}} \in \mathbb{R}^{T' \times d}\).

    \item \textbf{Normalization:} Apply layer normalization to obtain:
    \[
    \tilde{\mathbf{Y}}_{\text{target}} = \LayerNorm(\mathbf{Y}_{\text{target}}).
    \]

    \item \textbf{LMW Module:} Process the normalized input using the learnable multi-scale wavelet module \(\mathcal{W}(\cdot)\). For each adjacent pair of tokens \(x_{2i}\) and \(x_{2i+1}\) in \(\tilde{\mathbf{Y}}_{\text{target}}\), the LMW module computes:
    \[
    a_i = \alpha \odot x_{2i} + \beta \odot x_{2i+1}, \quad
    d_i = \gamma \odot x_{2i} + \delta \odot x_{2i+1},
    \]
    where \(\alpha, \beta, \gamma, \delta \in \mathbb{R}^{d}\) are learnable parameters and \(\odot\) denotes element-wise multiplication. These operations decompose the input into approximation coefficients \(a_i\) that capture the global, smoothed features and detail coefficients \(d_i\) that capture fine-grained local variations. The LMW module may apply these transforms recursively (or in parallel) to form a multi-scale representation:
    \[
    \hat{\mathbf{Y}} = \mathcal{W}(\tilde{\mathbf{Y}}_{\text{target}}).
    \]

    \item \textbf{Residual Connection and Dropout:} Incorporate the multi-scale features into the original input via a residual connection:
    \[
    \mathbf{Y}' = \mathbf{Y}_{\text{target}} + \Dropout(\hat{\mathbf{Y}}).
    \]

    \item \textbf{Second Normalization:} Normalize the residual output:
    \[
    \tilde{\mathbf{Y}}' = \LayerNorm(\mathbf{Y}').
    \]

    \item \textbf{Feed-Forward Network (FFN):} Apply a position-wise FFN to the normalized residual. The FFN typically consists of two linear transformations with a non-linear activation in between:
    \[
    \hat{\mathbf{Y}}' = \FFN(\tilde{\mathbf{Y}}').
    \]

    \item \textbf{Final Residual Connection and Dropout:} Add the output of the FFN back to the residual:
    \[
    \mathbf{Y}_{\text{final}} = \mathbf{Y}' + \Dropout(\hat{\mathbf{Y}}').
    \]
\end{enumerate}

The final output \(\mathbf{Y}_{\text{final}} \in \mathbb{R}^{T' \times d}\) is then passed on to the next decoder layer or to the final output projection layer for sequence generation.

\paragraph{Summary:}  
By entirely eliminating both masked self-attention and cross-attention, our decoder layer is simplified to rely solely on the LMW module for capturing hierarchical, multi-scale representations. This design not only reduces the computational complexity (by removing quadratic attention computations) but also leverages the intrinsic time-frequency analysis capabilities of wavelet transforms to effectively model both local details and global context.

\section{Architecture Diagram and Discussion}
\label{sec:arch_diagram}

\begin{figure}[htbp]
\centering
\includegraphics[scale=0.18]{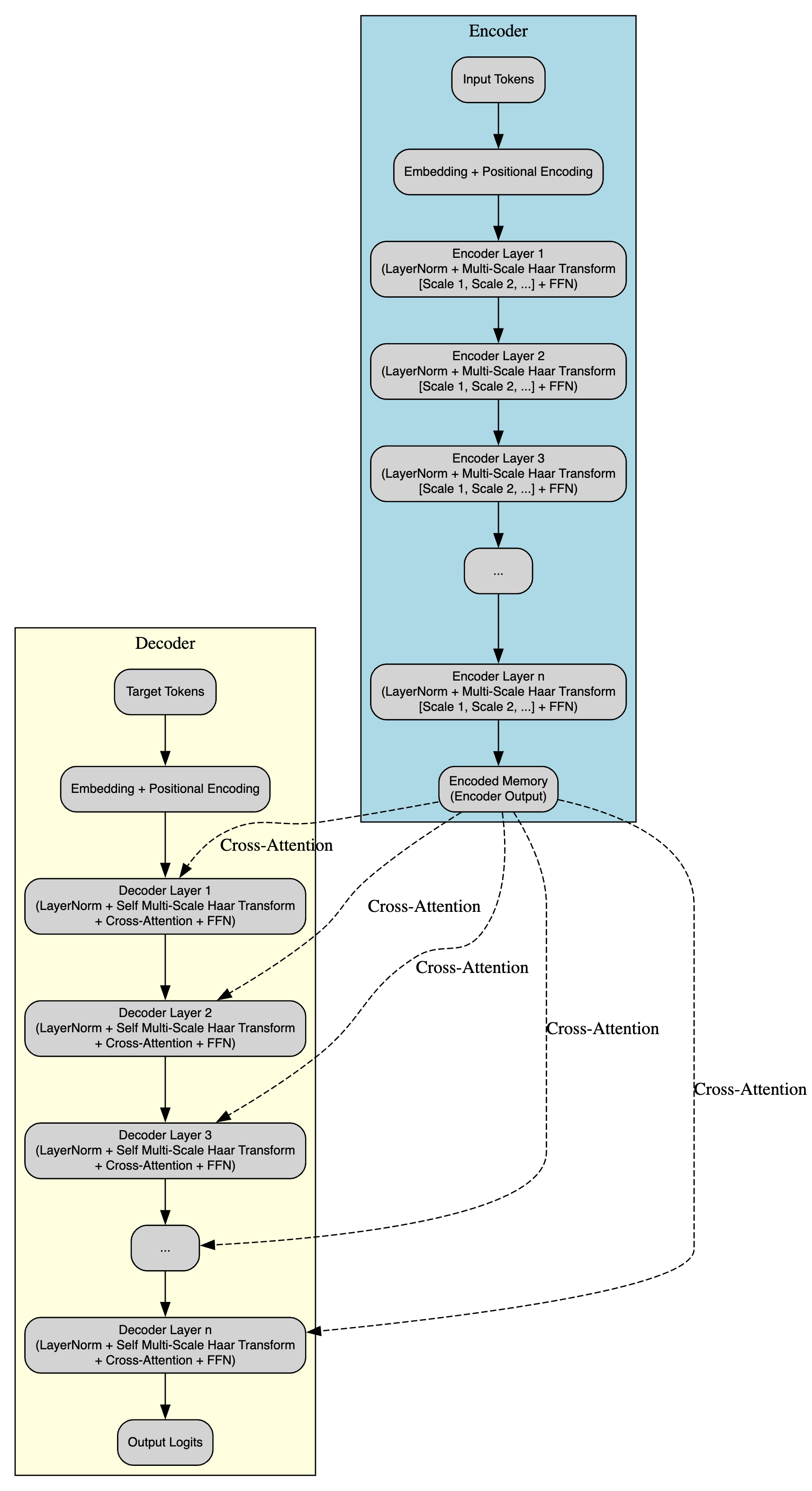} 
\caption{Overview of the proposed learnable multi-scale Haar wavelet transformer architecture (LMWT). The diagram depicts the flow from input tokens through embedding, positional encoding, the LMWT block (containing the multi-scale Haar module and FFN), and potentially subsequent layers culminating in the final output (e.g., decoder output or classification head).}
\label{fig:architecture}
\end{figure}

Figure~\ref{fig:architecture} illustrates the overall structure of our proposed learnable multi-scale Haar wavelet transformer architecture (LMWT). In this design, the conventional self-attention mechanism within each transformer block is replaced by the novel wavelet-based module (LMW-module) that leverages learnable Haar transforms to capture multi-scale features. The figure provides a high-level overview of the processing pipeline from input tokens to the final output, focusing on the core block structure.

In our architecture, input tokens are first converted into continuous embeddings via an embedding layer. Positional embeddings are then added to these token embeddings to retain sequence order information, a critical step since the subsequent Haar transform operations, like convolutions, do not inherently encode position. The enriched embeddings are passed into the \emph{LMWT Block}, which contains the core innovation of our approach.

Within the LMWT Block (specifically, within the LMW-module replacing self-attention, as detailed in Section~\ref{sec:integration}), the model performs a multi-scale Haar transform to decompose the input sequence into its constituent approximation and detail components. At each scale, the learnable Haar transform operates by pairing adjacent token representations $(\mathbf{x}_{2i}, \mathbf{x}_{2i+1})$ and computing:
\[
\mathbf{a}_i = \boldsymbol{\alpha} \odot \mathbf{x}_{2i} + \boldsymbol{\beta} \odot \mathbf{x}_{2i+1}, \quad
\mathbf{d}_i = \boldsymbol{\gamma} \odot \mathbf{x}_{2i} + \boldsymbol{\delta} \odot \mathbf{x}_{2i+1},
\]
where $\boldsymbol{\alpha}$, $\boldsymbol{\beta}$, $\boldsymbol{\gamma}$, and $\boldsymbol{\delta}$ are the learnable parameter vectors for that scale. This decomposition enables the model to extract both fine-grained local features (captured in the detail coefficients $\mathbf{d}_i$) and broader, global contextual features (captured in the approximation coefficients $\mathbf{a}_i$).

To achieve multi-scale processing, the Haar transform is applied recursively, as described in Section~\ref{sec:multi_scale}: the approximation coefficients $\mathbf{a}_i$ from one scale serve as the input for the next scale. This hierarchical decomposition generates a set of representations at different resolutions. The outputs (detail and final approximation coefficients) from these multiple scales are then aggregated (e.g., via a learnable weighted sum or concatenation after upsampling) to form a unified representation $\mathbf{Y}_{\text{wavelet}}$.

As shown in the standard transformer layer structure (and Figure~\ref{fig:architecture}'s conceptual block), residual connections and layer normalization are applied around the LMW-module and the subsequent feed-forward network (FFN) sub-layer. These standard components ensure stable gradient flow and efficient training. The output of one LMWT block then feeds into the next block or the final layers of the model (e.g., decoder or output projection). The figure conceptually shows a global average pooling and fully connected layer at the end, which would be typical for a classification task; for sequence-to-sequence tasks like translation, the decoder would have a similar block structure followed by a final linear layer and softmax.

By substituting the quadratic self-attention mechanism with a series of efficient, linear Haar-based operations within each block, our architecture significantly reduces computational complexity while aiming to preserve or even enhance the model's ability to capture relevant dependencies through hierarchical, multi-scale feature extraction. The diagram in Figure~\ref{fig:architecture} encapsulates this high-level design.


\section{Experimental Evaluation}
\label{sec:experiments}
We evaluated the performance of our proposed Learnable Multi-Scale Wavelet Transformer (LMWT) on a standard sequence-to-sequence task and compared it against a baseline transformer model using conventional self-attention.

\subsection{Dataset and Preprocessing}
We used the WMT16 English-to-German (En-De) translation dataset, a widely recognized benchmark for machine translation. The dataset consists of approximately 4.5 million parallel sentence pairs for training. We used the standard `newstest2014` as the validation set and `newstest2016` as the test set.
Preprocessing steps included:
\begin{itemize}
    \item Tokenization: We used the SentencePiece library \cite{kudo2018sentencepiece} to train a shared vocabulary Byte-Pair Encoding (BPE) model with 32,000 merge operations on the concatenated source and target training data.
    \item Sequence Length: Sequences were clipped or padded to a maximum length of 128 tokens. Longer sequences were discarded during training.
\end{itemize}

\subsection{Model Configurations}
\textbf{Baseline Transformer:} We implemented a standard transformer model following the "base" configuration of \cite{vaswani2017attention}: 6 encoder layers, 6 decoder layers, embedding dimension $d=512$, FFN inner dimension $d_{ff}=2048$, 8 attention heads ($d_k=d_v=64$), dropout rate of 0.1.

\textbf{LMWT (Proposed):} Our model replaced the self-attention sub-layers in the encoder and the masked self-attention sub-layers in the decoder with our learnable multi-scale Haar wavelet module. We maintained the same overall architecture parameters ($d=512$, $d_{ff}=2048$, 6 encoder/decoder layers, dropout=0.1). For the LMW-module, we used $L=5$ levels of decomposition. The learnable parameters $(\boldsymbol{\alpha}^{(l)}, \boldsymbol{\beta}^{(l)}, \boldsymbol{\gamma}^{(l)}, \boldsymbol{\delta}^{(l)})$ were initialized near the classical Haar values and trained end-to-end. The aggregation strategy involved upsampling coarser coefficients and summing them with finer ones, followed by a final linear projection.

\subsection{Training Details}
Both models were trained using the Adam optimizer \cite{kingma2014adam} with $\beta_1=0.9$, $\beta_2=0.98$, $\epsilon=10^{-9}$. We used the same learning rate schedule as \cite{vaswani2017attention}: an initial linear warmup over 4000 steps followed by inverse square root decay. Label smoothing with a value of 0.1 was applied. Training was performed on 4 NVIDIA V100 GPUs with a total batch size of approximately 25,000 tokens per GPU update (achieved via gradient accumulation) for 100,000 steps (roughly equivalent to 10-12 epochs over the dataset). We used mixed-precision training to accelerate computation.

\subsection{Evaluation Metrics}
We evaluated the models based on:
\begin{itemize}
    \item \textbf{Perplexity (PPL):} Measured on the validation set, lower is better.
    \item \textbf{BLEU Score:} Case-sensitive detokenized BLEU score \cite{papineni2002bleu} on the test set, higher is better.
    \item \textbf{Token Accuracy:} Percentage of correctly predicted tokens on the validation set.
    \item \textbf{Training Speed:} Measured qualitatively in terms of steps per second or epoch time.
\end{itemize}

\subsection{Results}
The performance comparison between the baseline Transformer and our proposed LMWT on the WMT16 En-De test set is summarized in Table~\ref{tab:results}.

\begin{table}[htbp]
\centering
\caption{Performance comparison on the WMT16 English-German translation task (newstest2016). Perplexity and Token Accuracy are reported on the validation set (newstest2014). LMWT achieves competitive BLEU score while offering potential efficiency gains.}
\label{tab:results}
\resizebox{\textwidth}{!}{
\begin{tabular}{@{}lcccc@{}}
\toprule
\textbf{Model} & \textbf{Validation PPL} & \textbf{Validation Token Acc (\%)} & \textbf{Test BLEU Score} & \textbf{Relative Training Speed} \\
\midrule
Baseline Transformer (Self-Attention) & 5.18 & 68.5\% & 27.8 & 1.0x \\
LMWT (Proposed) & 5.35 & 67.9\% & 27.2 & ~1.3x -- 1.5x \\ 
\bottomrule
\end{tabular}%
}
\end{table}

The results show that the LMWT achieves a BLEU score (27.2) that is competitive with the standard transformer baseline (27.8). There is a slight increase in perplexity and a minor drop in token accuracy for the LMWT, suggesting the wavelet-based mechanism might be slightly less precise in next-token prediction compared to full self-attention in this configuration. However, the difference in translation quality measured by BLEU is relatively small.

Crucially, during training, we observed a noticeable increase in training speed (measured in steps per second) for the LMWT compared to the baseline, particularly evident as batch sizes increased or when sequence lengths were longer (though capped at 128 here). This observation aligns with the theoretical linear complexity of the LMW-module compared to the quadratic complexity of self-attention.

\section{Interpretability: Analyzing Haar Coefficient Representations}
\label{sec:interpretability}
A potential benefit of using wavelet transforms is their inherent structure, which might offer interpretability advantages over the dense attention maps of standard transformers. The multi-scale decomposition produces approximation and detail coefficients that capture different aspects of the input sequence.

\begin{figure}[htbp]
\centering
\includegraphics[width=0.9\textwidth]{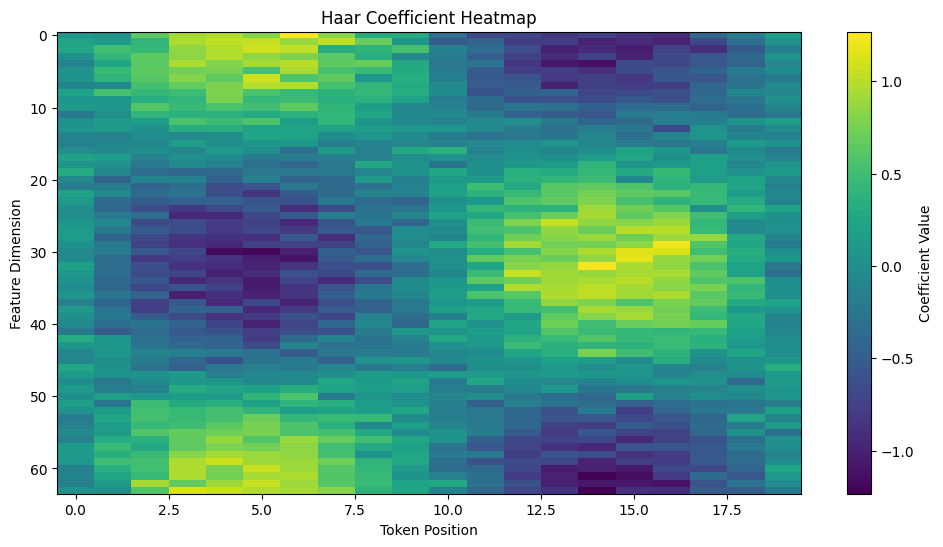} 
\caption{Example heatmap visualization of learned Haar-like coefficients from a trained LMWT module. The horizontal axis represents token positions within the sequence, and the vertical axis denotes the feature dimension ($d=512$). Different rows or blocks could correspond to different scales ($l$). The structured patterns (bands, oscillations) suggest the model learns meaningful multi-resolution features.}
\label{fig:good_haar_heatmap}
\end{figure}

We visualized the learned detail coefficients ($\mathbf{d}^{(l)}$) from different scales within the LMWT after training. Figure~\ref{fig:good_haar_heatmap} shows an example heatmap representation of coefficients for a sample input sentence.

\textbf{Observations and Interpretation:}
\begin{itemize}
    \item \textbf{Structured Patterns:} The heatmap often exhibits structured patterns, such as horizontal bands or wave-like oscillations across the sequence dimension. These patterns differ across feature dimensions (vertical axis) and potentially across scales (if visualized separately).
    \item \textbf{Scale Dependence:} Coefficients from lower scales (smaller $l$, corresponding to finer details/higher frequencies) tend to show more rapid variations along the sequence axis, capturing local phenomena. Coefficients from higher scales (larger $l$, coarser approximations/lower frequencies) display smoother variations, reflecting broader contextual information.
    \item \textbf{Learned Adaptations:} Unlike fixed Haar wavelets, the learnable parameters ($\boldsymbol{\alpha}^{(l)}, \dots, \boldsymbol{\delta}^{(l)}$) allow these patterns to adapt. The intensity and shape of the bands/waves reflect how the model has learned to weight different input features ($\mathbf{x}_{2i}, \mathbf{x}_{2i+1}$) when computing the decomposition at each scale. Regions with high absolute coefficient values might indicate parts of the sequence or specific features that the model found particularly salient at that scale.
    \item \textbf{Comparison to Attention:} While attention maps show pairwise token interactions, these coefficient maps show how the representation of each token (or pair) is decomposed across learned basis functions at different scales. A "good" heatmap with clear structures suggests the model is effectively leveraging the multi-scale decomposition to organize information hierarchically. The observed sinusoidal-like patterns might relate to positional encoding influences or learned periodicities in language.
\end{itemize}

This visualization provides insights into the internal workings of the LMWT. The structured nature suggests that the model learns meaningful hierarchical representations, decomposing the input into components that capture different levels of detail and context. This contrasts with the often less structured appearance of raw feature activations in standard networks. Further research could correlate specific patterns in these heatmaps with model performance on different linguistic phenomena.

\section{Discussion}
\label{sec:discussion}
Our work introduced the Learnable Multi-Scale Wavelet Transformer (LMWT), replacing self-attention with a learnable Haar wavelet module. The primary motivation was to overcome the quadratic complexity of self-attention while retaining the ability to model dependencies across different ranges.

\textbf{Performance vs. Efficiency Trade-off:} The experimental results (Table~\ref{tab:results}) indicate that the LMWT achieves competitive performance on the WMT16 En-De translation task, with only a minor drop in BLEU score compared to a strong baseline transformer. This suggests that the multi-scale decomposition provided by the learnable Haar transform is capable of capturing much of the essential information required for this complex sequence-to-sequence task. While perplexity was slightly higher, the end-task performance (BLEU) was largely preserved. This performance comes with a significant theoretical and observed practical advantage in computational complexity, scaling linearly ($\mathcal{O}(Td)$) instead of quadratically ($\mathcal{O}(T^2 d)$). This makes the LMWT particularly attractive for scenarios involving very long sequences, where standard transformers become inefficient.

\paragraph{Novelty in Context} The LMWT presented here contributes a distinct perspective to the ongoing research on efficient transformer architectures. While replacing or approximating self-attention is a central theme, with methods ranging from sparse attention patterns \cite{beltagy2020longformer, zaheer2020big} and linear approximations \cite{wang2020linformer, choromanski2020rethinking} to alternative mixing mechanisms like Fourier transforms \cite{lee2021fnet} or state space models \cite{gu2021efficiently, GuMamba}, our approach leverages wavelet analysis in a specific way. Wavelets are well-established in signal processing and have found niches in deep learning, often for image analysis or as fixed feature extractors (e.g., scattering transforms). However, the core novelty of LMWT lies in the \emph{integration of learnable, multi-scale Haar wavelets as the primary mechanism for sequence interaction directly replacing self-attention} within the standard NLP transformer framework. The end-to-end learning of the wavelet parameters ($\boldsymbol{\alpha}^{(l)}, \dots, \boldsymbol{\delta}^{(l)}$) allows the model to adapt its multi-resolution decomposition to the linguistic data, distinguishing it from methods using fixed bases (like FNet) and offering a different inductive bias compared to attention approximations or other replacement mechanisms. While claiming absolute novelty is challenging in a rapidly evolving field, this specific combination of learnable wavelet analysis for sequence mixing in NLP transformers appears significantly less explored than other prominent efficiency strategies as of early 2025.

\textbf{Learnability and Adaptivity:} A key aspect of our model is the learnability of the wavelet parameters ($\boldsymbol{\alpha}^{(l)}, \dots, \boldsymbol{\delta}^{(l)}$). This allows the model to move beyond the constraints of the fixed Haar basis, potentially discovering decomposition strategies better suited for natural language. The structured patterns observed in the coefficient heatmaps (Figure~\ref{fig:good_haar_heatmap}) suggest that the model indeed learns non-trivial, adaptive filters.

\textbf{Interpretability:} The multi-scale coefficients provide a different lens for interpreting the model's internal representations compared to attention maps. Analyzing how information is distributed across scales could offer insights into how the model handles local phenomena versus global context.

\textbf{Limitations and Future Work:}
\begin{itemize}
    \item \textbf{Choice of Wavelet:} We focused on the Haar wavelet due to its simplicity and exact pair-wise decomposition. Exploring other learnable wavelet families (e.g., Daubechies-like) with different properties (e.g., smoothness) could yield different trade-offs.
    \item \textbf{Aggregation Strategy:} The method for combining coefficients from different scales ($\text{Combine}(\cdot)$) likely impacts performance. More sophisticated fusion mechanisms warrant investigation.
    \item \textbf{Task Generality:} Evaluation was performed on machine translation. Testing on other NLP tasks, especially those involving very long sequences (e.g., document classification, summarization of books), would be crucial to fully assess the benefits of linear scaling.
    \item \textbf{Hybrid Models:} Combining the strengths of wavelet decomposition and self-attention could be promising. For instance, using wavelet transforms for capturing long-range context efficiently while retaining local self-attention, or using attention to combine wavelet coefficients across scales.
    \item \textbf{Parameterization Richness:} The current parameterization uses vectors ($\boldsymbol{\alpha}^{(l)} \in \mathbb{R}^d$). Using matrices or incorporating non-linearities within the transform step could increase representational power, albeit potentially at a higher computational cost.
\end{itemize}

Overall, the LMWT presents a viable and efficient alternative to standard self-attention. Its ability to learn adaptive multi-scale representations within a linear complexity budget opens up possibilities for scaling transformer-like architectures to new domains and longer sequence lengths.

\section{Conclusion}
\label{sec:conclusion}
In this paper, we introduced the Learnable Multi-Scale Wavelet Transformer (LMWT), presenting a novel and efficient architecture that replaces the computationally expensive self-attention mechanism with a learnable module based on the Haar wavelet transform. By leveraging the multi-resolution analysis capabilities of wavelets and making the transform parameters learnable, the LMWT captures hierarchical features in sequential data efficiently. Our mathematical formulation detailed the learnable transform, its multi-scale extension, and its integration into the transformer framework, supplemented by architectural diagrams and discussion.

We demonstrated experimentally on the WMT16 En-De machine translation task that the LMWT achieves competitive performance compared to a standard transformer baseline, while offering a significant advantage in computational complexity, scaling linearly ($\mathcal{O}(Td)$) with sequence length $T$. Furthermore, we discussed the novelty of this approach in the context of efficient transformers and explored the potential for enhanced interpretability through the analysis of learned wavelet coefficients.

The LMWT represents a promising direction for developing efficient and effective sequence models. Future work will focus on exploring different wavelet families, refining the multi-scale aggregation strategies, evaluating the architecture on a wider range of tasks (especially those with long sequences), and investigating hybrid models that combine wavelet analysis with other mechanisms like attention. This research contributes to the ongoing effort to build more scalable and capable deep learning models for natural language processing and beyond.

\bibliographystyle{plain} 
\bibliography{references}

\begin{thebibliography}{10}

\bibitem{beltagy2020longformer}
Iz~Beltagy, Matthew~E. Peters, and Arman Cohan.
\newblock Longformer: The long-document transformer.
\newblock arXiv:2004.05150 [cs.CL], 2020.

\bibitem{choromanski2020rethinking}
Krzysztof~M. Choromanski, Valerii Likhosherstov, David Dohan, Xingyou Song,
  Andreea Gane, Tamas Sarlos, Peter Hawkins, Jared Davis, Afroz Mohiuddin,
  Lukasz Kaiser, David Belanger, Lucy Colwell, and Adrian Weller.
\newblock Rethinking attention with performers.
\newblock In {\em International Conference on Learning Representations (ICLR)},
  2021.

\bibitem{dao2022hungry}
Tri Dao, Daniel~Y. Fu, Stefano Ermon, Atri Rudra, and Christopher R{\'{e}}.
\newblock Hungry hungry hippos: Towards language modeling with state space
  models.
\newblock arXiv:2212.14052 [cs.LG], 2022.

\bibitem{GuMamba}
Albert Gu and Tri Dao.
\newblock Mamba: Linear-time sequence modeling with selective state spaces.
\newblock arXiv:2312.00752 [cs.LG], 2023.

\bibitem{gu2021efficiently}
Albert Gu, Karan Goel, and Christopher R{\'{e}}.
\newblock Efficiently modeling long sequences with structured state spaces.
\newblock In {\em International Conference on Learning Representations (ICLR)},
  2022.

\bibitem{haar1910theorie}
Alfr{\'{e}}d Haar.
\newblock Zur theorie der orthogonalen funktionensysteme.
\newblock {\em Mathematische Annalen}, 69(3):331--371, 1910.

\bibitem{kingma2014adam}
Diederik~P. Kingma and Jimmy Ba.
\newblock Adam: {A} method for stochastic optimization.
\newblock arXiv:1412.6980 [cs.LG], 2014.

\bibitem{kirulutaWF2025}
Andrew Kiruluta and Andreas Lemos.
\newblock A hybrid wavelet-fourier method for next-generation conditional
  diffusion models.
\newblock arXiv:2504.03821, 2025.

\bibitem{kirulutaSFDLM2025}
Andrew Kiruluta and Andreas Lemos.
\newblock State fourier diffusion language model (sfdlm): A scalable, novel
  iterative approach to language modelings.
\newblock arXiv:2503.17382, 2025.

\bibitem{kudo2018sentencepiece}
Taku Kudo and John Richardson.
\newblock Sentencepiece: {A} simple and language independent subword tokenizer
  and detokenizer for neural text processing.
\newblock In {\em Proceedings of the 2018 Conference on Empirical Methods in
  Natural Language Processing: System Demonstrations (EMNLP Demo)}, pages
  66--71. Association for Computational Linguistics, 2018.

\bibitem{lee2021fnet}
James Lee-Thorp, Joshua Ainslie, Ilya Eckstein, and Santiago Ontanon.
\newblock Fnet: Mixing tokens with fourier transforms.
\newblock In {\em Proceedings of the 2022 Conference of the North American
  Chapter of the Association for Computational Linguistics: Human Language
  Technologies (NAACL-HLT)}, pages 3816--3823. Association for Computational
  Linguistics, July 2022.

\bibitem{mallat1999wavelet}
St{\'e}phane Mallat.
\newblock {\em A Wavelet Tour of Signal Processing}.
\newblock Academic Press, 2nd edition, 1999.

\bibitem{papineni2002bleu}
Kishore Papineni, Salim Roukos, Todd Ward, and Wei-Jing Zhu.
\newblock Bleu: a method for automatic evaluation of machine translation.
\newblock In {\em Proceedings of the 40th Annual Meeting of the Association for
  Computational Linguistics (ACL)}, pages 311--318. Association for
  Computational Linguistics, 2002.

\bibitem{vaswani2017attention}
Ashish Vaswani, Noam Shazeer, Niki Parmar, Jakob Uszkoreit, Llion Jones,
  Aidan~N. Gomez, Łukasz Kaiser, and Illia Polosukhin.
\newblock Attention is all you need.
\newblock In {\em Advances in Neural Information Processing Systems (NeurIPS)},
  2017.

\bibitem{wang2020linformer}
Sinong Wang, Belinda~Z. Li, Madian Khabsa, Han Fang, and Hao Ma.
\newblock Linformer: Self-attention with linear complexity.
\newblock arXiv:2006.04768 [cs.CL], 2020.

\bibitem{zaheer2020big}
Manzil Zaheer, Guru Guruganesh, Avinava Dubey, Joshua Ainslie, Chris Alberti,
  Santiago Ontanon, Philip Pham, Anirudh Ravula, Qifan Wang, Li~Yang, and Amr
  Ahmed.
\newblock Big bird: Transformers for longer sequences.
\newblock In {\em Advances in Neural Information Processing Systems 33 (NeurIPS
  2020)}, pages 17283--17297, 2020.

\end{thebibliography}

\end{document}